\documentclass[11pt]{article}

\usepackage[preprint]{acl}

\usepackage{times}
\usepackage{latexsym}
\usepackage{multirow}

\usepackage[T1]{fontenc}

\usepackage{microtype}

\usepackage{inconsolata}

\usepackage{graphicx}
\graphicspath{{figs/}}
\usepackage{amsmath}
\usepackage{algorithm}
\usepackage{algpseudocode}

\title{SCOPE: Structured Decomposition and Conditional Skill Orchestration for Complex Image Generation}
\author{
  \bfseries
  Tianfei Ren\textsuperscript{1},
  Zhipeng Yan\textsuperscript{1},
  Yiming Zhao\textsuperscript{1},
  Zhen Fang\textsuperscript{1*},
  Yu Zeng\textsuperscript{1*},
\\
  \bfseries
  Guohui Zhang\textsuperscript{1},
  Hang Xu\textsuperscript{1},
  Xiaoxiao Ma\textsuperscript{1},
  Shiting Huang\textsuperscript{1},
  Ke Xu\textsuperscript{1},
  Wenxuan Huang,
\\
  \bfseries
  Lionel Z. Wang\textsuperscript{2,3},
  Lin Chen\textsuperscript{1},
  Zehui Chen\textsuperscript{1},
  Jie Huang\textsuperscript{1},
  Feng Zhao\textsuperscript{1\ensuremath{\dagger}}
\\
\\
  \textsuperscript{1}MoE Key Laboratory of Brain-inspired Intelligent Perception and Cognition,\\
  University of Science and Technology of China
\\
  \textsuperscript{2}The Hong Kong Polytechnic University \quad
  \textsuperscript{3}Nanyang Technological University
\\
  \textsuperscript{*}Project lead. \quad
  \textsuperscript{\ensuremath{\dagger}}Corresponding author.
\\[0.8em]
{\small Project page: \url{https://nopnor.github.io/SCOPE/}}
\\[1.0em]
\includegraphics[width=0.98\textwidth]{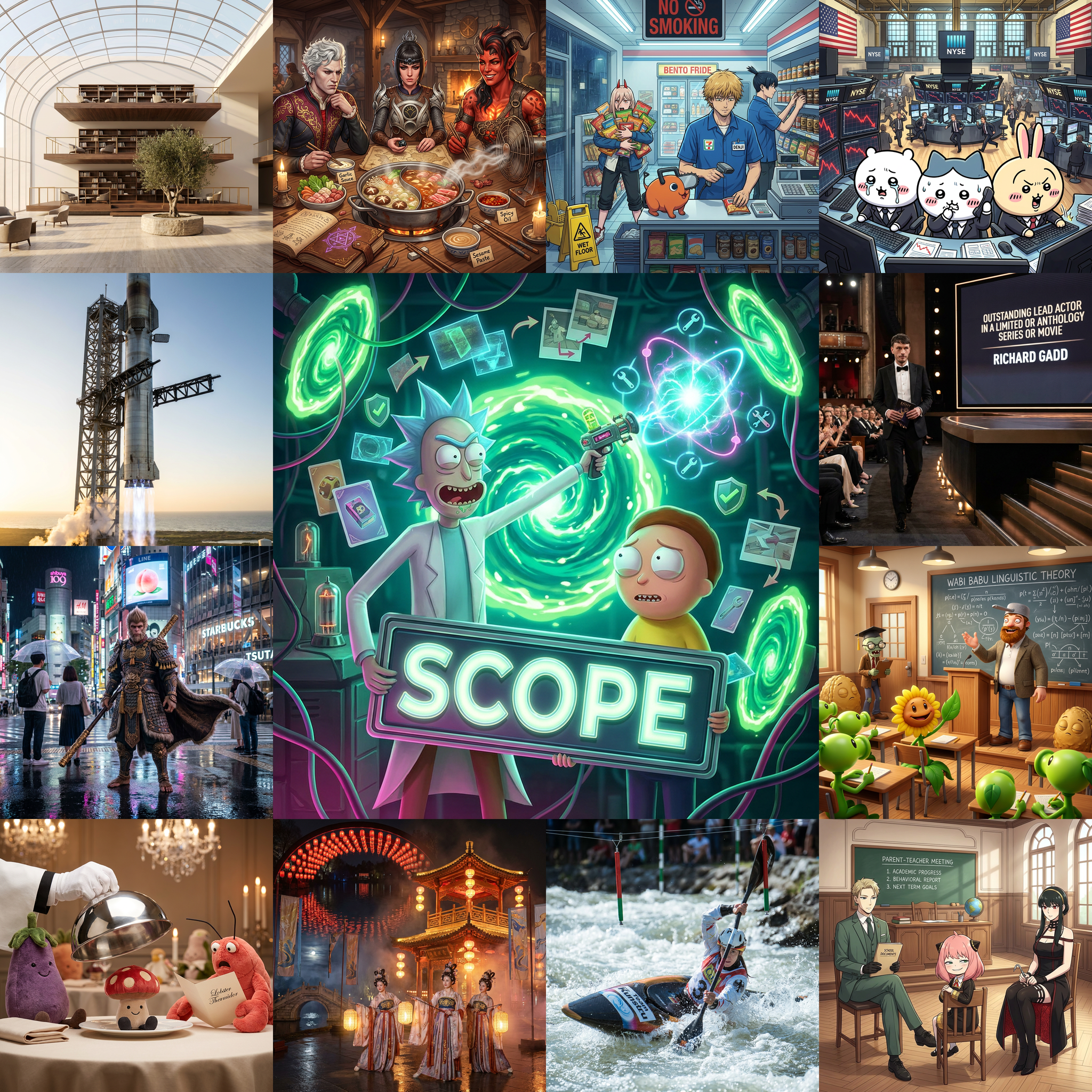}
\\[0.45em]
\refstepcounter{figure}\label{fig:teaser}%
\parbox{0.94\textwidth}{\raggedright\small
\textbf{Figure~\thefigure:} Examples generated by SCOPE across knowledge-intensive events, reference-heavy intellectual properties, and multi-entity compositions. SCOPE maintains structured commitments and invokes skills to resolve or repair
them throughout generation, leading to \textbf{SOTA} performance on Gen-Arena and strong results on external benchmarks.
}
}

\begin{document}
\maketitle
\clearpage
\begin{abstract}
While text-to-image models have made strong progress in visual fidelity, faithfully realizing complex visual intents remains challenging because many requirements must be tracked across grounding, generation, and verification. We refer to these requirements as \textbf{semantic commitments} and formalize their lifecycle discontinuity as the \textit{Conceptual Rift}, where commitments may be locally resolved or checked but fail to remain identifiable as the same operational units throughout the generation lifecycle. To address this, we propose \textbf{SCOPE}, a specification-guided skill orchestration framework that maintains semantic commitments in an evolving structured specification and conditionally invokes retrieval, reasoning, and repair skills around unresolved or violated commitments. To evaluate commitment-level intent realization, we introduce \textbf{Gen-Arena}, a human-annotated benchmark with entity- and constraint-level specifications, together with \textbf{Entity-Gated Intent Pass Rate (EGIP)}, a strict entity-first pass criterion. SCOPE substantially outperforms all evaluated baselines on Gen-Arena, achieving 0.60 EGIP, and further achieves strong results on WISE-V (0.907) and MindBench (0.61), demonstrating the effectiveness of persistent commitment tracking for complex image generation.
\end{abstract}

\section{Introduction}
Text-to-image generation is shifting from producing plausible images toward realizing user-specified visual intent. Earlier progress was often measured by whether generated images looked convincing and aligned with prompts at a coarse level~\cite{saharia2022photorealistic,betker2023improving,esser2024scaling}. As these systems enter practical and creative workflows, however, prompts increasingly describe scenes whose defining details must hold for the result to be faithful~\cite{hu2023tifa,huang2023t2i,ghosh2023geneval,li2024evaluating}. This changes the standard for generation: an image should not only appear natural, but also correctly instantiate the scene the user intended to depict.

Faithfully realizing complex visual intents is challenging because the requirements imposed by a prompt do not all become actionable at the same time. Some requirements are explicit from the beginning, while others become determinate only after the system grounds external context or reasons about what the scene implies~\cite{li2025ia,son2025world,he2026mind,feng2026gen,chen2026unify}. We refer to these requirements as \textbf{semantic commitments}: conditions that the final image must satisfy for the user's intent to be fulfilled. The challenge is therefore not only to discover these commitments, but also to keep each one identifiable until it can be visually realized and checked.

Recent work has increasingly moved beyond one-shot prompt-to-image generation by introducing multi-step interventions such as retrieval, planning, and iterative refinement~\cite{li2025ia,son2025world,feng2026gen, chen2025t2i,li2025reflect,wuvisualprompter, ye2025genpilot}. These interventions improve different aspects of complex generation, from resolving missing information to correcting visible failures. However, making generation multi-step does not by itself ensure lifecycle continuity: a commitment may be resolved before generation, checked after generation, and revised in a later step, yet these operations may not remain tied to the same identifiable unit. As a result, even when a system retrieves relevant information or identifies a real failure, the generation target may still be diluted, the error may be misattributed, or the repair action may be poorly targeted. We refer to this lifecycle discontinuity as the \textbf{Conceptual Rift}. This raises a central question: \textbf{\textit{how can semantic commitments remain representable, verifiable, and actionable as unified operational units throughout the generation lifecycle?}}

To address this challenge, we propose \textbf{SCOPE}, a specification-guided skill orchestration framework for complex image generation. SCOPE makes semantic commitments explicit in an evolving structured specification, which serves as a shared interface across the generation lifecycle. Guided by this specification, SCOPE conditionally invokes retrieval, reasoning, and repair skills to ground missing external information, infer implicit requirements, and revise violated commitments. It also verifies generated images against the commitments represented in the current specification. By writing skill outputs and verification results back to the specification, SCOPE keeps complex visual intents actionable across stages, allowing generation to proceed through a unified lifecycle rather than a sequence of disconnected local operations.

If complex generation is organized around semantic commitments, evaluation should also reveal which commitments are fulfilled or violated. Existing evaluations often rely on holistic alignment scores, while checklist-style protocols may still treat conditions as largely independent, obscuring whether a failure is a root error or a downstream consequence of missing prerequisite content. We introduce \textbf{Gen-Arena}, an entity- and constraint-level benchmark that represents each prompt with a structured evaluation specification. By linking constraints to their prerequisite entities, Gen-Arena supports entity-first evaluation and enables diagnosis of missing entities, violated constraints, and downstream failures caused by unmet prerequisites.

In summary, our contributions are as follows:
\begin{itemize}
    \item We identify the \textbf{Conceptual Rift} in complex image generation, where semantic commitments behind a complex visual intent lose continuity across the generation lifecycle, and propose \textbf{SCOPE}, which addresses this rift by maintaining these commitments in an evolving specification and orchestrating retrieval, reasoning, and repair skills around it. 
    \item We introduce \textbf{Gen-Arena}, a human-annotated benchmark for commitment-level intent realization, with entity- and constraint-level evaluation specifications and \textbf{Entity-Gated Intent Pass Rate (EGIP)} as a strict entity-first pass criterion.
    \item Experiments show that SCOPE substantially outperforms all evaluated baselines on Gen-Arena, achieving 0.60 EGIP, and further achieves strong results on WISE-V (0.907) and MindBench (0.61), demonstrating the value of persistent commitment tracking for complex image generation.
\end{itemize}

\section{Related Work}
\subsection{Agentic Image Generation}
Recent work has begun to use multimodal agents to mediate complex image generation beyond direct prompt conditioning~\cite{chen2025t2i,ye2025genpilot,xiang2025promptsculptor,wang2025imagent,garg2026sidiffagent,he2026gems,han2026unicorn}. Existing methods improve this process from different angles: some strengthen the interpretation of user requests~\cite{chen2025t2i,ye2025genpilot,xiang2025promptsculptor}, some ground generation or multimodal reasoning with retrieved visual or factual evidence~\cite{li2025ia,son2025world,he2026mind,feng2026gen,chen2026unify,huang2026vision,zeng2026vision}, and others refine outputs through reflection or feedback~\cite{li2025reflect,zhuo2025reflection,wuvisualprompter,jaiswal2026iterative,venkatesh2025crea,huang2025interleaving}. These approaches demonstrate the value of agentic mediation for complex generation. However, their intermediate representations are usually tailored to particular interventions, rather than to maintaining the same semantic commitments across the full generation lifecycle. As a result, resolved information, verification outcomes, and repair decisions may not remain reliably tied to the same underlying commitments.

\subsection{Skills in Language and Multimodal Agents}
Prior work broadly views skills as reusable procedural knowledge that extends language agents beyond one-off tool use~\cite{liu2024skillact,jiang2026sok,li2026skillsbench}. Such skills may be written by humans, distilled from demonstrations or trajectories, or selected from large repositories according to the current task~\cite{zheng2025skillweaver,zheng2026skillrouter}. Recent multimodal systems further show that skill abstractions can support complex visual reasoning and generation workflows: XSkill~\cite{jiang2026xskill} accumulates task-level skills from visual-tool trajectories, while GEMS~\cite{he2026gems} introduces memory and domain skills for agent-native multimodal generation. However, most existing work treats skills primarily as reusable agent resources. Less explored is how skill use should be grounded in the evolving semantic commitments of a specific task, so that each invocation addresses a concrete unresolved or violated item and its result remains usable in later stages.

\section{Method}
\label{sec:method}

We first introduce the overall design of SCOPE in Section~\ref{sec:scope_framework}. We then describe how the evolving structured specification supports conditional skill orchestration. Finally, we introduce the construction of Gen-Arena in Section~\ref{sec:gen_arena}.

\subsection{SCOPE Framework}
\label{sec:scope_framework}

\subsubsection{Overall Pipeline}

\begin{figure*}[t]
\centering
\includegraphics[width=0.95\textwidth]{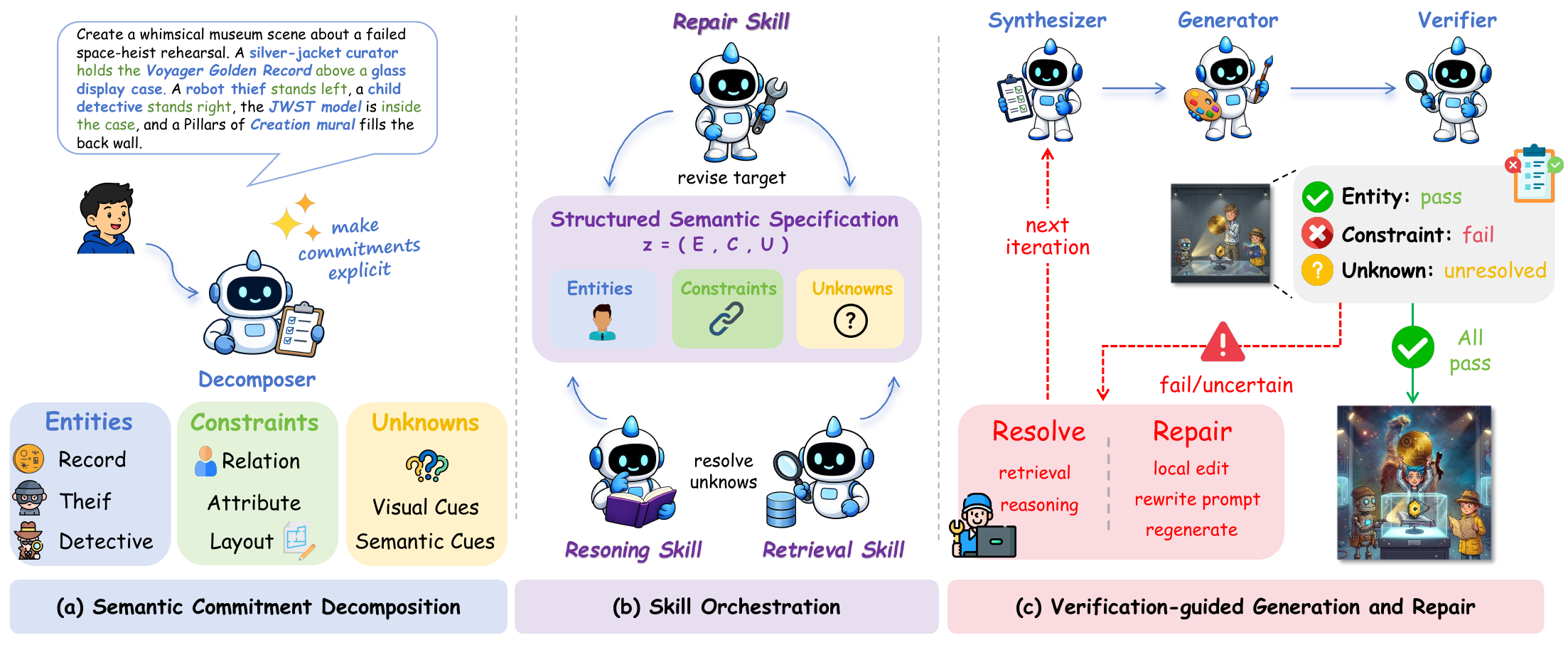}
\caption{Overview of SCOPE. The user prompt is decomposed into an evolving structured semantic specification containing entities, constraints, and unknowns. Retrieval, reasoning, and repair skills update this specification around unresolved or violated commitments, while verification maps generated images back to the same commitment-level representation.}
\label{fig:framework}
\end{figure*}

SCOPE is designed to keep semantic commitments operational across the generation lifecycle. It represents the current visual intent as an evolving structured specification and uses this specification as the shared interface for generation, verification, and targeted skill invocation. Given a user prompt and an optional reference image, SCOPE iteratively operates over a fixed core pipeline of \textbf{Decomposer $\rightarrow$ Synthesizer $\rightarrow$ Generator $\rightarrow$ Verifier}. Specifically, the \textbf{Decomposer} transforms the user request into the specification, the \textbf{Synthesizer} consolidates resolved information from the current specification into a coherent generation prompt, the \textbf{Generator} performs image generation or editing, and the \textbf{Verifier} evaluates entities and constraints item by item. In addition to this fixed core loop, SCOPE conditionally orchestrates retrieval, reasoning, and repair skills to address unresolved semantics and localized generation failures as they arise. Figure~\ref{fig:framework} shows the overall SCOPE architecture, where the structured specification provides the shared interface for generation, verification, and targeted skill invocation.

\subsubsection{Structured Semantic Specification}

To keep semantic commitments identifiable across the generation lifecycle, SCOPE represents each request as an evolving structured semantic specification $z=(E,C,U)$. Here, $E$ denotes the target entities that should be instantiated in the image, $C$ denotes verifiable commitments over these entities, and $U$ denotes unresolved information that prevents reliable realization of a commitment. We group constraints in $C$ into three types. Attribute constraints specify entity-level requirements such as identity, appearance, quantity, and visible text. Relation constraints specify interactions or semantic relationships between entities. Layout constraints specify the placement of entities within the scene and their composition with the surrounding environment. Importantly, unknowns in $U$ are not treated as independent questions. Each unknown is attached to a prompt-, entity-, or constraint-level owner, indicating which commitment it is meant to resolve. This ownership link allows retrieval or reasoning results to update the corresponding part of the specification, and later allows verification failures to be mapped back to the same entities or constraints. When a verified failure is no longer associated with unresolved unknowns, the mapped entity or constraint becomes the target of repair. Thus, the specification provides the operational interface for carrying resolved information, verification outcomes, and repair targets across iterations.

\subsubsection{Conditional Skill Orchestration}
SCOPE orchestrates retrieval, reasoning, and repair skills through the current specification rather than applying them as fixed pipeline steps. Retrieval is invoked when a commitment depends on missing external evidence, such as factual information or reference identity cues. Reasoning is invoked when an implicit or underspecified commitment must be resolved before reliable synthesis. Repair is invoked when verification maps a visual failure back to an entity or constraint that no longer requires additional grounding or reasoning. Each skill is anchored to a concrete unknown or violated commitment, and its output updates the same specification: retrieval and reasoning close unresolved unknowns, while repair records how violated items are revised. In this way, SCOPE adapts the generation lifecycle according to what remains unresolved or violated in the current specification.

\subsubsection{Verification-Guided Resolution and Repair}
Given the current output image $y_t$ at iteration $t$ and the specification $z=(E,C,U)$, SCOPE verifies the image against the explicit item set $I=E\cup C$, rather than relying on a holistic image-level judgment. The Verifier returns itemized review results $R_t=\{r_i^{(t)}\}_{i\in I}$, where each $r_i^{(t)}$ records a verdict $v_i^{(t)}\in\{\texttt{pass},\texttt{fail},\texttt{uncertain}\}$ and a textual reason. We define the set of items requiring further action as
\begin{equation}
F_t=\{\,i\in I \mid v_i^{(t)}\in\{\texttt{fail},\texttt{uncertain}\}\,\}.
\end{equation}

For each item in $F_t$, SCOPE maps the verification result back to the current specification. If the item is associated with an unresolved or newly exposed unknown, the issue is treated as a remaining semantic gap, and retrieval or reasoning is invoked to continue resolution. Otherwise, the issue is treated as a visual realization failure: the commitment is already specified, but the generated image does not satisfy it. In this case, SCOPE invokes repair on the violated entity or constraint.

The repair skill selects among prompt rewriting, image editing, and regeneration according to the scope of the failure. Prompt rewriting is used when the synthesized prompt does not faithfully express the current specification, image editing is used for localized defects, and regeneration is used when the failure is broad or too entangled for reliable local correction. Thus, verification serves as the routing mechanism between continued semantic resolution and targeted visual repair, keeping post-generation actions grounded in the same specification used before generation. Algorithm~\ref{alg:scope} summarizes how SCOPE operates across the generation lifecycle.

\begin{algorithm}[t]
\caption{SCOPE Generation Lifecycle}
\label{alg:scope}
\small
\begin{algorithmic}[1]
\Require User prompt $p$, maximum attempts $T$
\State $z_0=(E_0,C_0,U_0) \leftarrow \Call{Decompose}{p}$
\For{$t=1$ to $T$}
    \State $z_t \leftarrow z_{t-1}$
    \ForAll{$u \in U_t$ that remains unresolved}
        \If{$u$ requires external evidence}
            \State $e_u \leftarrow \Call{Retrieve}{u}$
        \Else
            \State $e_u \leftarrow \Call{Reason}{u,z_t}$
        \EndIf
        \State $z_t \leftarrow \Call{UpdateSpec}{z_t,u,e_u}$
    \EndFor
    \State $s_t \leftarrow \Call{SynthesizePrompt}{z_t}$
    \State $y_t \leftarrow \Call{Generate}{s_t}$
    \State $R_t \leftarrow \Call{Verify}{y_t,z_t}$
    \If{\Call{AllPass}{$R_t$}}
        \State \Return $y_t$
    \EndIf
    \State $F_t \leftarrow \Call{MapFailures}{R_t,z_t}$
    \ForAll{$i \in F_t$}
        \If{$i$ exposes an unresolved unknown}
            \State $z_t \leftarrow \Call{AddUnknown}{z_t,i}$
        \Else
            \State $z_t \leftarrow \Call{Repair}{z_t,i,y_t}$
        \EndIf
    \EndFor
\EndFor
\State \Return best verified image $y^\ast$
\end{algorithmic}
\end{algorithm}

\subsection{Gen-Arena}
\label{sec:gen_arena}
Gen-Arena evaluates whether complex image generation fulfills structured visual intents rather than only matching prompts at a coarse level. Each instance pairs a natural-language prompt with an evaluation specification that identifies the required entities and the constraints they must satisfy, enabling commitment-level evaluation of generated images.

\begin{figure*}[t]
\centering
\includegraphics[width=0.98\textwidth]{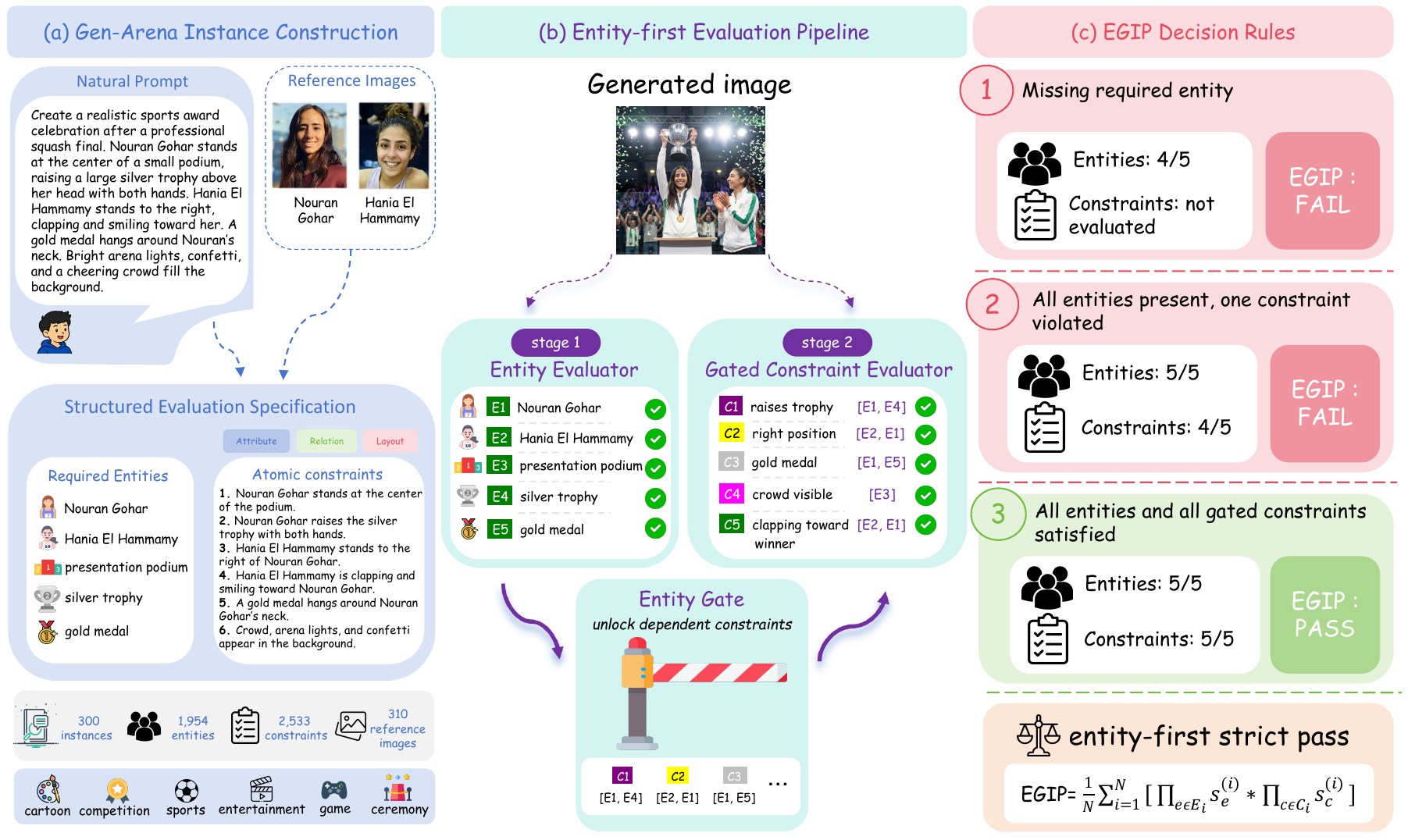}
\caption{Overview of Gen-Arena construction and EGIP evaluation. Gen-Arena represents each prompt with required entities and atomic constraints, links constraints to prerequisite entities, and evaluates generation with an entity-first pass rule.}
\label{fig:gen_arena}
\end{figure*}

Gen-Arena is manually constructed through a human annotation pipeline covering six categories: cartoon, game, sports, entertainment, competition, and ceremony. Annotators first write natural user prompts and collect reference images when identity or appearance cannot be specified reliably by text alone. They then identify visible target entities in each prompt and annotate atomic constraints over these targets, including attributes, relations, and layouts. Each constraint is linked to the entities it depends on, allowing evaluation to distinguish missing-entity failures from unsatisfied constraints over correctly realized entities. The resulting benchmark contains 300 instances, 1,954 entities, 2,533 constraints, and 310 reference images. Figure~\ref{fig:gen_arena} summarizes the Gen-Arena construction pipeline and the entity-gated evaluation protocol.

For evaluation, Gen-Arena uses an entity-first strict pass rule. The evaluator first checks whether all required entities are correctly realized. If any required entity is missing or incorrectly depicted, the instance is marked as failed. Only when all required entities are satisfied does the evaluator check the associated constraints. Let $s_e^{(i)}\in\{0,1\}$ indicate whether entity $e\in E_i$ is correctly realized, and let $s_c^{(i)}\in\{0,1\}$ indicate whether constraint $c\in C_i$ is satisfied. We define \textbf{Entity-Gated Intent Pass Rate (EGIP)} as
\begin{equation}
    \mathrm{EGIP}
    =
    \frac{1}{N}\sum_{i=1}^{N}
    \left(
    \prod_{e \in E_i} s_e^{(i)}
    \prod_{c \in C_i} s_c^{(i)}
    \right).
\end{equation}
Thus, EGIP measures strict instance-level intent fulfillment: an example passes only when all required entities and constraints are satisfied.


\section{Experiments}
We evaluate SCOPE from two perspectives: whether it improves commitment-level realization on Gen-Arena, and whether the same framework transfers to existing complex generation benchmarks.
\subsection{Experimental Setup}

\paragraph{Baselines and Benchmarks.}
We evaluate SCOPE on Gen-Arena and two external benchmarks: WISE-V~\cite{niu2025wise} and MindBench from Mind-Brush~\cite{he2026mind}. Gen-Arena uses EGIP to measure strict commitment-level intent fulfillment across six categories, while the external benchmarks test transfer to existing evaluations involving world knowledge and reasoning-intensive visual generation. On Gen-Arena, we compare against closed-source models Nano Banana~\cite{google-deepmind-2025-nano-banana} and Nano Banana Pro~\cite{google-deepmind-2025-nano-banana-pro}, as well as open-source models SDXL~\cite{podell2023sdxl}, SD-3.5-large~\cite{esser2024scaling}, FLUX.1-dev~\cite{black-forest-labs-2024-flux}, Qwen-Image~\cite{wu2025qwen}, Z-Image-Turbo~\cite{cai2025z}, PixArt-Sigma~\cite{chen2024pixart}, and Janus-Pro-7B~\cite{chen2025janus}.

\begin{table*}[t]
\centering
\small
\setlength{\tabcolsep}{3pt}
\renewcommand{\arraystretch}{1.15}
\begin{tabular}{lccccccc}
\hline
Method & Cartoon & Game & Sports & Ent. & Comp. & Ceremony & Overall \\
\hline
Nano Banana~\cite{google-deepmind-2025-nano-banana} & 0.02 & 0.00 & 0.16 & 0.14 & 0.10 & 0.00 & 0.07 \\
Nano Banana Pro~\cite{google-deepmind-2025-nano-banana-pro} & 0.04 & 0.00 & 0.34 & 0.16 & 0.18 & 0.54 & 0.21 \\
\hline
Qwen-Image~\cite{wu2025qwen} & 0.00 & 0.00 & 0.10 & 0.02 & 0.00 & 0.00 & 0.02 \\
Z-Image-Turbo~\cite{cai2025z} & 0.00 & 0.00 & 0.09 & 0.02 & 0.00 & 0.00 & 0.01 \\
PixArt-Sigma~\cite{chen2024pixart} & 0.00 & 0.00 & 0.00 & 0.00 & 0.00 & 0.00 & 0.00 \\
FLUX.1-dev~\cite{black-forest-labs-2024-flux} & 0.00 & 0.00 & 0.03 & 0.02 & 0.00 & 0.00 & 0.01 \\
SD-3.5-large~\cite{esser2024scaling} & 0.00 & 0.00 & 0.00 & 0.02 & 0.00 & 0.00 & 0.00 \\
SDXL~\cite{podell2023sdxl} & 0.00 & 0.00 & 0.00 & 0.00 & 0.00 & 0.00 & 0.00 \\
Janus-Pro-7B~\cite{chen2025janus} & 0.00 & 0.00 & 0.00 & 0.00 & 0.00 & 0.00 & 0.00 \\
SCOPE & 0.52 & 0.46 & 0.72 & 0.62 & 0.52 & 0.74 & 0.60 \\
\hline
\end{tabular}
\caption{Main results on Gen-Arena. We report Entity-Gated Intent Pass Rate (EGIP) for each category and the overall benchmark. Ent., Comp. abbreviate entertainment and competition, respectively.}
\label{tab:gen_arena_main}
\end{table*}

\begin{table*}[t]
\centering
\small
\setlength{\tabcolsep}{2.6pt}
\renewcommand{\arraystretch}{1.15}
\resizebox{\textwidth}{!}{%
\begin{tabular}{lccccccc|ccc}
\hline
Method
& \multicolumn{7}{c|}{WISE-V}
& \multicolumn{3}{c}{MindBench} \\
& Culture & Time & Space & Biology & Physics & Chemistry & Overall
& Knowledge & Reasoning & Overall \\
\hline
Nano Banana Pro~\cite{google-deepmind-2025-nano-banana-pro} & \textbf{0.898} & 0.817 & 0.933 & 0.817 & 0.867 & 0.875 & 0.876
& 0.40 & 0.44 & 0.41 \\
GPT-Image-1.5~\cite{openai-2025-gpt-image-15} & 0.890 & 0.692 & 0.883 & 0.800 & 0.758 & 0.775 & 0.825
& 0.22 & 0.18 & 0.21 \\
\hline
Bagel w/ CoT~\cite{deng2025emerging} & 0.780 & 0.633 & 0.567 & 0.375 & 0.550 & 0.508 & 0.628
& - & - & - \\
Qwen-Image~\cite{wu2025qwen} & 0.628 & 0.525 & 0.558 & 0.342 & 0.483 & 0.250 & 0.510
& 0.02 & 0.02 & 0.02 \\
Z-Image~\cite{cai2025z} & 0.548 & 0.467 & 0.508 & 0.325 & 0.475 & 0.175 & 0.453
& 0.02 & 0.00 & 0.02 \\
FLUX.1-dev~\cite{black-forest-labs-2024-flux} & 0.523 & 0.400 & 0.533 & 0.175 & 0.375 & 0.242 & 0.416
& 0.01 & 0.03 & 0.02 \\
FLUX.1-schnell~\cite{black-forest-labs-2024-flux} & 0.465 & 0.325 & 0.467 & 0.208 & 0.383 & 0.100 & 0.364
& - & - & - \\
SD-3.5-large~\cite{esser2024scaling} & 0.490 & 0.408 & 0.442 & 0.300 & 0.375 & 0.208 & 0.404
& 0.02 & 0.03 & 0.01 \\
SD-3.5-medium~\cite{esser2024scaling} & 0.483 & 0.375 & 0.375 & 0.183 & 0.392 & 0.200 & 0.376
& 0.01 & 0.00 & 0.01 \\
SDXL~\cite{podell2023sdxl} & 0.493 & 0.367 & 0.242 & 0.267 & 0.333 & 0.183 & 0.364
& 0.02 & 0.01 & 0.01 \\
Bagel~\cite{deng2025emerging} & 0.413 & 0.350 & 0.308 & 0.200 & 0.442 & 0.258 & 0.352
& 0.00 & 0.03 & 0.02 \\
Janus-Pro-7B~\cite{chen2025janus} & 0.370 & 0.350 & 0.283 & 0.283 & 0.400 & 0.233 & 0.334
& - & - & - \\
Mind-Brush~\cite{he2026mind} & - & - & -
& - & - & - & -
& 0.38 & 0.24 & 0.31 \\
SCOPE & 0.883 & \textbf{1.000} & \textbf{0.944} & \textbf{0.861} & \textbf{0.889} & \textbf{0.917} & \textbf{0.907}
& \textbf{0.59} & \textbf{0.63} & \textbf{0.61} \\
\hline
\end{tabular}
}
\caption{External benchmark results on WISE-V~\cite{niu2025wise} and MindBench from Mind-Brush~\cite{he2026mind}. WISE-V reports category-level WiScore averages and the official overall score; MindBench reports representative submetrics and overall accuracy.}
\label{tab:external_benchmarks}
\end{table*}

\paragraph{Implementation Details.}
We use GPT-5.4 as the MLLM backend and Nano Banana Pro as the image generation and editing backend. Retrieval is implemented with Google Search API, and the maximum number of generation attempts is set to 3 for each case. For Gen-Arena, Gemini 3-Pro serves as the official evaluator, judging entities and constraints item by item. When an entity is associated with reference images, the evaluator is instructed to compare the generated image against the provided references rather than relying only on the text prompt. 


\subsection{Main Results}

We report main results in two parts. Table~\ref{tab:gen_arena_main} evaluates methods on Gen-Arena with commitment-level metrics, while Table~\ref{tab:external_benchmarks} summarizes results on external benchmarks.

\begin{figure*}[t]
\centering
\includegraphics[width=0.98\textwidth]{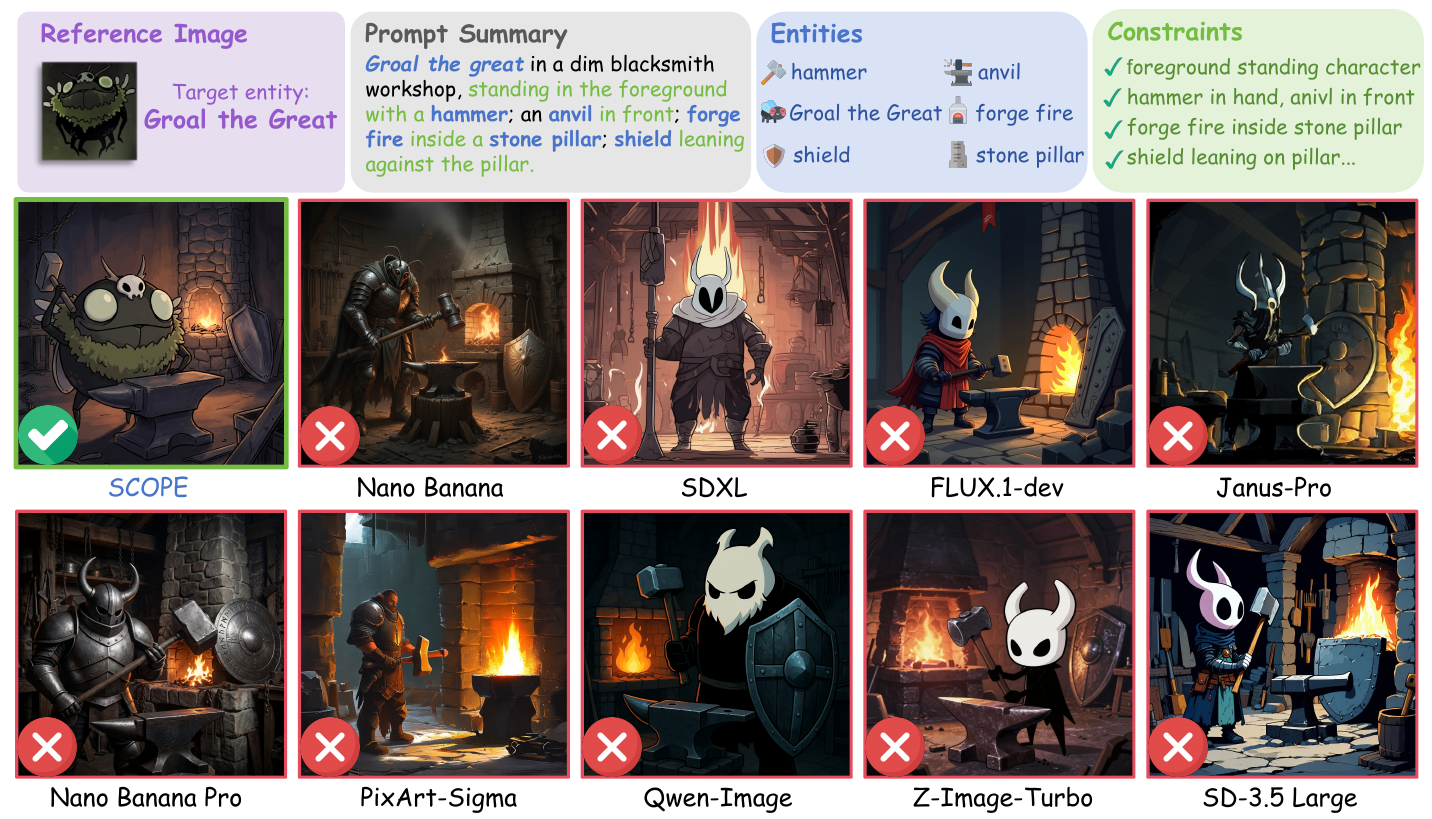}
\caption{Illustrative qualitative comparisons between direct prompting and SCOPE-guided generation using the same image backend. Examples cover entity/count, attribute/identity, relation, and layout failures.}
\label{fig:qualitative_compare}
\end{figure*}

Table~\ref{tab:gen_arena_main} presents the quantitative comparison on Gen-Arena. SCOPE achieves a significant improvement in overall EGIP compared to direct generation baselines, surpassing Nano Banana Pro by 39 percentage points. The improvement is consistent across all six categories, with especially strong results on Sports and Ceremony. These categories frequently require identity grounding, event-specific relations, and precise scene composition, suggesting that SCOPE benefits from keeping retrieved evidence, inferred requirements, and verification feedback tied to the same structured specification. In contrast, direct generation baselines often fail under the strict pass criterion even when individual visual elements appear plausible, showing that stronger generation alone is insufficient for complete structured intent realization. 
Figure~\ref{fig:qualitative_compare} compares generated images from baseline models and SCOPE on a representative Gen-Arena prompt.

As shown in Table~\ref{tab:external_benchmarks}, on WISE-V, SCOPE achieves the best overall WiScore of 0.907 and ranks first in five of the six reported categories, improving over Nano Banana Pro by 3.5\% overall. On MindBench, SCOPE achieves 0.63 in Reasoning and 0.61 overall, improving over Nano Banana Pro by 48.8\% in overall accuracy. These results further corroborate that SCOPE effectively maintains explicit commitments across the generation lifecycle, thereby realizing more faithful complex image generation. 

\subsection{Ablation Study}

\begin{table}[t]
\centering
\small
\setlength{\tabcolsep}{5.5pt}
\renewcommand{\arraystretch}{1.18}
\begin{tabular}{lcc}
\hline
Method & EGIP & Gated Cons. \\
\hline
Direct (single) & 0.21 & 0.59 \\
Direct (best-of-3) & 0.40 & 0.71 \\
Self-refine w/o spec & 0.39 & 0.73 \\
\hline
SCOPE w/o R\&R & 0.22 & 0.60 \\
SCOPE w/o repair & 0.42 & 0.72 \\
SCOPE & \textbf{0.60} & \textbf{0.83} \\
\hline
\end{tabular}
\caption{Ablation results on Gen-Arena. Direct uses the Nano Banana Pro backend with the original prompt; best-of-3 selects the best result among three independent generations using the official evaluator. Self-refine w/o spec uses the same three-generation budget but replaces structured commitments with free-form critique and prompt rewriting. R\&R denotes retrieval and reasoning, and Gated Cons. denotes gated constraint pass rate.}
\label{tab:ablation}
\end{table}

Table~\ref{tab:ablation} studies the contribution of the main SCOPE skills on Gen-Arena. Direct (single) obtains 0.21 EGIP, while Direct (best-of-3) rises to 0.40 by reporting the best official-evaluator score among three independent direct generations. Self-refine w/o spec reaches only 0.39 EGIP despite using the same three-generation budget, suggesting that free-form critique and prompt rewriting do not reliably convert partial improvements into complete instance-level success. Within SCOPE, removing retrieval and reasoning skills drops EGIP to 0.22, close to Direct (single), indicating that structured decomposition alone is insufficient unless exposed commitments are semantically resolved. Removing only repair skills reaches 0.42 EGIP, showing that retrieval and reasoning provide substantial initial-generation gains by resolving underspecified commitments. Full SCOPE reaches 0.60 EGIP, improving over w/o repair by 18 percentage points. These results suggest that SCOPE benefits from coupling a persistent structured specification with complementary skills: retrieval and reasoning resolve underspecified commitments before generation, while verification-guided repair targets commitments that remain violated later in the generation lifecycle.
\begin{table*}[t]
\centering
\small
\setlength{\tabcolsep}{3pt}
\renewcommand{\arraystretch}{1.1}
\begin{tabular}{lccc}
\hline
Method & EGIP & Entity Pass & Gated Constraint Pass \\
\hline
Nano Banana~\cite{google-deepmind-2025-nano-banana} & 0.07 & 0.78 & 0.49 \\
Nano Banana Pro~\cite{google-deepmind-2025-nano-banana-pro} & 0.21 & 0.82 & 0.59 \\
\hline
Qwen-Image~\cite{wu2025qwen} & 0.02 & 0.83 & 0.49 \\
Z-Image-Turbo~\cite{cai2025z} & 0.01 & 0.84 & 0.48 \\
PixArt-Sigma~\cite{chen2024pixart} & 0.00 & 0.67 & 0.31 \\
FLUX.1-dev~\cite{black-forest-labs-2024-flux} & 0.01 & 0.78 & 0.42 \\
SD-3.5-large~\cite{esser2024scaling} & 0.00 & 0.76 & 0.39 \\
SDXL~\cite{podell2023sdxl} & 0.00 & 0.56 & 0.23 \\
Janus-Pro-7B~\cite{chen2025janus} & 0.00 & 0.78 & 0.40 \\
SCOPE & 0.60 & 0.92 & 0.83 \\
\hline
\end{tabular}
\caption{Diagnostic Gen-Arena results. Entity Pass measures item-level entity satisfaction, while Gated Constraint Pass measures constraint satisfaction after applying entity prerequisites.}
\label{tab:gen_arena_diagnostic}
\end{table*}

\subsection{Analysis}
We further analyze SCOPE by asking two questions about the generation lifecycle.

\noindent\textbf{Q1: Is structured decomposition alone sufficient?}
The ablation results show that the structured specification should not be understood as a stronger prompt by itself. Its main function is to preserve addressable semantic units across stages. Without retrieval and reasoning, unresolved unknowns remain exposed in the specification; without repair, violations detected after generation cannot be targeted. Thus, the specification is useful because it provides a stable interface through which different skills can operate on the same commitments.

\noindent\textbf{Q2: Why does SCOPE improve EGIP?}
Table~\ref{tab:gen_arena_diagnostic} helps explain where SCOPE gains come from. Direct generation baselines pass many individual entity checks, but EGIP requires every required entity in an instance to be correct before its constraints can be credited. Thus, even an item-level Entity Pass of 0.82 for Nano Banana Pro still leaves many instances with at least one missing or incorrectly grounded entity, which then blocks dependent attributes, relations, or layouts. Qwen-Image and Z-Image-Turbo show a similar pattern: their Entity Pass exceeds 0.83, yet their EGIP remains near zero because entity errors and downstream constraint violations accumulate at the instance level. SCOPE improves EGIP by keeping entities and their associated constraints tied to the same structured specification across the generation lifecycle. This reduces cascading entity-to-constraint failures, raising Entity Pass to 0.92, Gated Constraint Pass to 0.83, and overall EGIP to 0.60.
\section{Conclusion}
We introduced SCOPE, a specification-guided skill orchestration framework for complex image generation. SCOPE addresses the Conceptual Rift by representing the semantic commitments imposed by complex visual intents in an evolving structured specification. It then orchestrates retrieval, reasoning, and repair skills around commitments that remain unresolved or violated. We further introduced Gen-Arena, a human-annotated benchmark for commitment-level evaluation, together with EGIP as a strict entity-first pass metric. Experiments show that SCOPE substantially outperforms all evaluated baselines on Gen-Arena and achieves strong results on external benchmarks, suggesting that keeping commitments identifiable across the generation lifecycle is a practical path toward more faithful complex image generation.

\section*{Limitations}

While SCOPE improves commitment-level intent realization, it also has several limitations. First, SCOPE requires multiple calls to the MLLM, image generator, verifier, and optionally retrieval services across up to three iterations per instance, leading to higher inference-time cost and latency than one-shot generation. This overhead could be reduced through adaptive early stopping or more selective skill invocation. Second, SCOPE depends on item-level verification to route subsequent repair actions. Verification errors can therefore propagate through the lifecycle: a false negative may trigger unnecessary repair of a correctly realized commitment, while a false positive may leave a genuine failure unresolved. Improving verifier calibration is therefore important for making verification-guided repair more reliable.

\bibliography{custom}

\end{document}